# Compressed DenseNet for Lightweight Character Recognition

**Zhao Zhang [1,2], Zemin Tang [2], Yang Wang [1], Haijun Zhang [3], Shuicheng Yan [4], Meng Wang [1]**

[1] Hefei University of Technology,  [2] Soochow University,  [3] Harbin Institute of Technology

[4] National University of Singapore

## Abstract

*Convolutional Recurrent Neural Network (CRNN) is a popular network for recognizing texts in images. Advances like the variant of CRNN, such as Dense Convolutional Network with Connectionist Temporal Classification, has reduced the running time of the network, but exposing the inner computation cost and weight size of the convolutional networks as a bottleneck. Specifically, the DenseNet based models utilize the dense blocks as the core module, but the inner features are combined in the form of concatenation in dense blocks. As such, the number of channels of combined features delivered as the input of the layers close to the output and the relevant computational cost grows rapidly with the dense blocks getting deeper. This will severely bring heavy computational cost and big weight size, which restrict the depth of dense blocks. In this paper, we propose a compressed convolution block called Lightweight Dense Block (LDB). To reduce the computing cost and weight size, we re-define and re-design the way of combining internal features of the dense blocks. LDB is a convolutional block similarly as dense block, but it can reduce the computation cost and weight size to (1/L, 2/L), compared with original ones, where L is the number of layers in blocks. Moreover, LDB can be used to replace the original dense block in any DenseNet based models. Based on the LDBs, we propose a Compressed DenseNet (CDenseNet) for the lightweight character recognition. Extensive experiments demonstrate that CDenseNet can effectively reduce the weight size while delivering the promising recognition results.*

## 1. Introduction

Texts and images are two of the most popular vision data in the area of computer vision. It is common in practice that texts are always embedded in images, so how to detect and recognize the texts or characters in images accurately by a learning algorithm is still challenging and an important topic in the field of visual pattern recognition [36, 37], such as the optical character recognition (OCR) [6]. OCR is a long-standing topic but still a very challenging task due to complicated background and complex contents in images [2]. Recent years have witness the fast development and continuous breakthroughs in computer vision [12-14] and deep learning [7-11][46][52], lots of advanced end-to-end deep learning methods have been proposed [24].

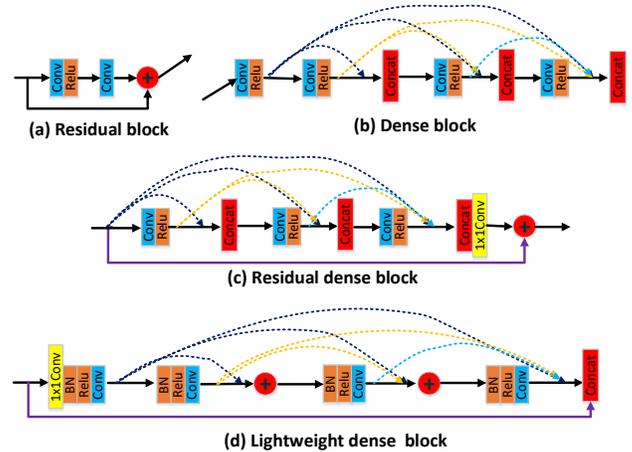

Figure 1. The structures of (a) residual block, (b) dense block, (c) residual dense block, and (d) our lightweight dense block.

For OCR, two crucial sub-tasks are text line extraction and text line recognition. The first task is to extract the regions of texts in images and the second one recognizes the textual contents of the identified region [24]. To handle the OCR, there are two mainstream frameworks at present. The first one is to train an end-to-end network that jointly solves the tasks of text line extraction and recognition, such as arbitrary orientation network (AON) [2]. The other popular one is a two-stage scheme, i.e., training two networks for two sub-tasks, for instance Convolutional Recurrent Neural Network (CRNN) [29]. Generally, the unified models have stronger adaptability and faster speed, but the results are slightly lower. The two-stage models usually have higher accuracy but lower efficiency. Recent CRNN integrates the advantages of Convolutional Neural Networks (CNNs) and Recurrent Neural Networks (RNNs) [18-20]. Since CRNN aims at recognizing the texts, it needs to extract the text lines in images by the novel Connectionist Text Proposal Network (CTPN) [4]. Recent work [51] revealed that even without the recurrent layers, the simplified model can still achieve promising result with higher efficiency. As such, the framework of CNN + CTC

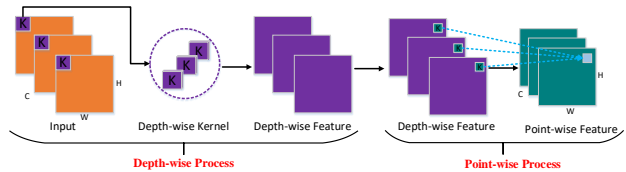

Figure 2. The process of depth separable convolution.

[1] is a feasible and efficient solution. To extract the features from convolutional layers, existing networks can be used, e.g., Dense Convolutional Network (DenseNet) [5] and Residual Network (ResNet) [16]. If DenseNet is used as the feature extractor and the recurrent layers are dropped for efficiency, one can form a new character recognition model, termed DenseNet + CTC.

It is noteworthy that DenseNet and its variants use the dense blocks as the core module to construct models for feature learning. However, the standard dense blocks have several drawbacks that may decrease its efficiency directly. First, to ensure the maximum information flow between different layers and to preserve the feed-forward nature, the feature maps of all preceding layers are used as the input of the current layer, and the feature maps of current layer are treated as the inputs of all subsequent layers in the standard dense blocks. However, this structure will bring about huge computing efforts and big weight size with deeper blocks, which will severely restrict the network depth and result in large amount of parameters. Besides, the pooling operation in DenseNet is unlearnable and may discard some important information. The traditional convolution operation used in DenseNet will also increase the amount of parameters.

In this paper, we therefore investigate the lightweight character recognition task via a compressed DenseNet, by refining the structures of blocks to reduce the weight size and cost. The contributions are summarized as follows:

- Technically, we propose a new compressed dense block termed Lightweight Dense Block (LDB) and then derive a lightweight dense convolutional network based on the designed LDBs. Specifically, we redefine and redesign the way of combining internal features over the original dense block to enable a more efficient block by weight-compression. Compared with the dense block, LDB uses both sum and concatenating operations to connect inner features in each block, which can reduce the computing cost and weight size to $(1/L, 2/L)$, where $L$ is the number of layers in blocks. Figure 1(d) illustrates the layout of our lightweight dense block, where the convolutional layer includes the functions of Batch Normalization (BN) [22], ReLU and Convolution. We mainly evaluate the designed LDBs to reduce the weight size, while maintaining the character recognition performance.

- We propose a lightweight character recognition network, termed Compressed DenseNet with Up-sampling block

(shortly, CDenseNet-U). We construct an up-sampling block using several designed LDBs and deconvolution operation for up-sampling features. To avoid important feature information loss and make the related parameters learnable, we use the convolution operation with stride 2 to replace the pooling operation. To further reduce the weight size and improve the efficiency of extracting dense features, we use the depth-wise separable convolution rather than original convolution. Moreover, we design two different convolutional operation groups for weight compression. One is shown in Figure 1(d), which is by adding a convolution with kernel size of 1*1, and with a scaling factor $t$ to reduce the channels of the input that usually results in big weight size, based on another convolutional group without 1*1 convolution.

- We conduct the character recognition tasks on the Chinese string and handwritten digits, which verifies that our proposed network can effectively reduce the weight size, while maintaining the recognition performance.

## 2. Related Work

### 2.1. Depth-wise Separable Convolution

The depth separable convolution, which is a new variant of the original convolution, is first presented in MobileNet [17]. Different from the original convolution that considers the channel and region changes jointly, the deep separable convolution clearly separates the channel and region and divides the convolution operation into two sub-steps, that is, depth-wise and point-wise processes.

The depth-wise process divides the input features with the form of $N \times H \times W \times C$ into $C$ groups and then performs convolution operation to each group, where $N$ is the number of features, $H$ and $W$ are the height and width of features, and $C$ is the number of channels of features. The depth-wise process collects spatial features of each channel, namely, depth-wise features. The point-wise mainly refers to a process that conducts $1 \times 1$ convolution operation using k filters to the output features from the depth-wise process, which collects the features from each point, i.e., point-wise features. Figure 2 shows the depth separable convolution in the case of padding "same", where K is convolution kernel.

### 2.2. Convolutional Recurrent Neural Network

CRNN is an end to end learning framework recognizing the text sequences in scenes [29]. CRNN takes advantage of the CNNs for the local feature extraction and RNN for the temporal summarization of extracted features. As shown in Figure 3, it consists of three parts, the first part denotes the Convolution layer that extracts the feature sequences from images, the second is the Recurrent Layer that predicts the label distribution of each frame, and the last layer is the transcription layer that transforms the prediction of each frame into the final label sequences.

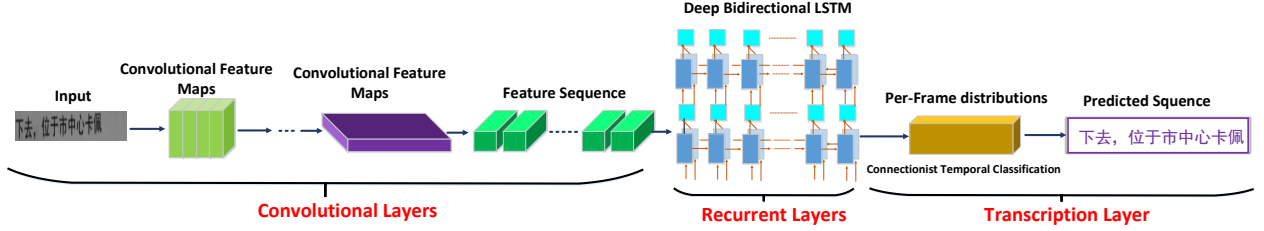

Figure 3. The learning architecture of CRNN for text recognition.

## 3. Proposed Lightweight Dense Block (LDB)

We describe the definition and structure of LDB, which is an important part of our lightweight character recognition framework. Since our LDB is inspired by the residual block [16], dense block [5] and residual dense block [50], we first review them and illustrate their structures in Figure 4 for comparison. Then, we detail the structure of our LDB.

### 3.1. Reviews of Related Blocks

**Residual block.** The Residual Network (ResNet) [16] was mainly proposed for addressing the issue of network degradation. The core idea of ResNet is embodied in the design of residual block. Figure 1(a) shows the structure of the standard residual block. Let $x$, $F(x)$ and $H(x)$ represent the input features, output features without and with a short connection respectively, we can obtain

$$F(x) = w_2\left(\sigma\left(w_1 x\right)\right),$$
$$H(x) = F(x) + x,$$  (1)

where $w_1$ and $w_2$ are the weights of convolutional layers, $\sigma(\bullet)$ is the function of Rectified Liner Units (ReLU) [23]. The identical maps can be constructed in two cases that either $F(x) = x$ without short connection or $F(x) = H(x) - x = 0$ with short connection. The optimization with a short connection is easier in training. The short connection in the residual block can also help enhance the feature flow.

**Dense block.** Different from ResNet, DenseNet mainly creates short paths between layers, and in this process a simple connectivity pattern, referred to as dense block [5], is derived. To ensure maximum information flow between layers, all layers with matching feature-map sizes in a dense block are connected directly with each other [5], and the related features are combined by concatenating them as

$$X_i = \phi\left(\left[X_0, X_1, ..., X_{i-1}\right]\right),$$  (2)

where $X_i$ represents the features of $i$-th layer, $X_0$ represents the input features and $\phi(\bullet)$ denotes the according function of convolution layer. To preserve the feed-forward nature, each layer obtains additional inputs from preceding layers and passes on its own feature-maps to subsequent layers [5]. Figure 1(b) illustrates the layout of dense block. DenseNet uses many short connections to enhance feature flow, so it has a strong feature learning ability. But the approach of combining the features by concatenating them will bring about sharp increase in terms of number of feature channels and huge computing efforts with the dense block getting deeper, which will restrict the depth of the networks using dense blocks, for instance DenseNet, directly.

**Residual dense block.** The residual dense block (RDB) is proposed in the residual dense network (RDN) [50] that aims to make full use of all hierarchical features from the original image. The core idea of RDN is to utilize both residual and dense blocks to improve the feature learning ability by enhancing the information flow, learning residual features and enhancing the local feature fusion. Figure 1(c) illustrates the structure of RDB. It should be noted that RDB utilizes the concatenating operation as dense block to combine the features of the former RDBs and current layers, and use the residual outside the dense block to enhance the representation learning. As such, the proposed RDB will also suffer from huge computing cost as the dense block. But its structural design gives us inspiration.

### 3.2. Structure of Lightweight Dense Block (LDB)

The main idea of our LDB uses both the sum operation in residual block and the concatenating operation in the dense block. But the difference is that RDB mainly improves the of feature representation ability by using the sum operation to construct a residual outside the dense block, while LDB mainly uses the sum operation to change the feature fusion mode inside the dense block such that the computing cost and weight size of the dense blocks can be clearly reduced. Specifically, LDB is proposed by redefining and designing the way of combining the internal features of the dense blocks. Generally, the number of channels of input features is more than that of the inner layers in dense block. As such, we propose to reduce the computing cost and weight size by using the sum operation to combine features instead of concatenating for all inner layers of dense blocks, except the input and output layer, which is performed as follows:

$$X_1 = \phi\left(X_0\right)$$
$$X_i = \phi\left(X_1 + ... + X_{i-1}\right), \ i > 1$$  (3)

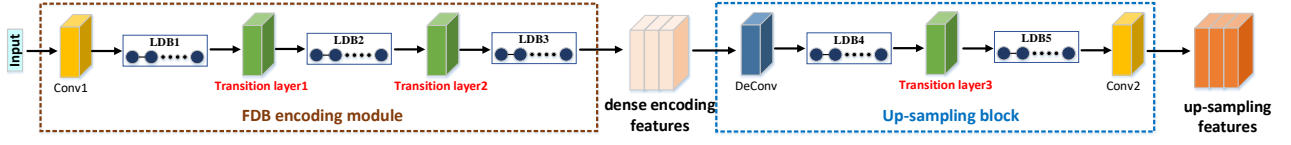

Figure 4. The structure of the convolutional network part of our proposed CDenseNet-U framework.

We obtain the output features by concatenating all the features of different layers in LDBs. As such, we can fully utilize feature information of all layers and can also ensure that our feature maps have the same size and same number of channels as that of dense block. As such, our proposed LDB will be applicable to any deep network structures that the original dense block can be used.

### 3.3. Theoretical Computational Analysis of LDB

As claimed above, LDB can reduce the computing cost and weight size of the dense block by redefining and designing the way of combining internal features, so we would like to present the theoretical computational analysis. Specifically, we summarize the analysis in Theorem 1.

**Theorem 1.** LDB can reduce the computing cost and weight size to ($1/L$, $2/L$), compared with the original dense block, where $L$ is the number of layers in the block.

**Proof.** Recalling the computation of combining features in the original dense blocks and our LDBs in formulas (2) and (3). For the original dense block, the input features of the $i$-th layer are the combined features of its all preceding layers by concatenating. For our LDB, the input feature of the first convolution layer is the original data and the input features of the $i$-th layer are the combined features of its preceding layers by summing over the $1$-th to ($i$-$1$)-th layers. Thus, we find that the number of channels of input features grows in the original dense blocks but keeps unchanged in the middle layers of our proposed LDB. Next, we present a quantitative comparison of the computational cost between the original dense block and our LDB.

Denote by $M$, $N$, $L$, $H$, $S$ the number of channels of input data, the number of kernels in each layer, the number of the convolutional layers in a dense block/our proposed LDB, the computing cost of each channel of each kernel and the weight size of each channel of each kernel, respectively. By setting the related parameters of the two blocks to be same, we can obtain the following computing efforts for original dense block and our proposed LDB:

$$
\begin{aligned}
C_{Dense} &= C_1 + C_2 + ... + C_l \\
&= H*\left(N*M + N*(M+N) + ... + N*(M+N*(L-1))\right), \quad (4) \\
&= H*N*\left(M*L + N*L*(L-1)/2\right)
\end{aligned}
$$

$$
\begin{aligned}
C_{LDB} &= C_1 + C_2 + ... + C_l \\
&= H*\left(N*M + N*N + ... + N*N\right), \quad (5) \\
&= H*N*\left(M + N*(L-1)\right)
\end{aligned}
$$

where $C_{Dense}$, $C_{FDB}$ and $C_i$ ($i = 1, 2,...l$) denote the computing efforts of the original dense block, LDB and each layer of the two blocks, respectively. Then, we have the following relation between $C_{Dense}$ and $C_{LDB}$:

$$
\frac{C_{LDB}}{C_{Dense}} = \frac{1+(1/\vartheta)}{L}, \ where\ \vartheta = 1 + \frac{2M}{N*(L-1)}. \quad (6)
$$

We can find that: when the value of $M$ is fixed, if the values of $L$ and $N$ are very large, the reduced computational effort tends to $2/L$, else the reduced cost tends to $1/L$. As such, one can easily conclude that $C_{LDB} / C_{Dense} \in (1/L, 2/L)$. Clearly, LDB gets an obvious computing effort reduction. It should be noted that the calculation method of the weight size is similar to that of the computational cost, and we can obtain a similar result, i.e., our LDB can clearly reduce the weight size compared with the original dense block. More specifically, by comparing with the dense block, we can draw the following conclusion that the computation and weight size of our LDB can be reduced to nearly $2/L$ with the increase of $L$ and $N$, but the value of $L$ is usually large at this time; otherwise, the cost reduction is about $1/L$, but the $L$ is smaller. This also fully demonstrates the rationality of our proposed LDB. Based on the computational analysis, reducing the cost of calculation and weight size have a clear restriction with LDB getting deeper, i.e., the reduced cost cannot be increased indefinitely with the growth of $L$ and $N$. Specifically, with the increasing number of the convolution layers inside LDB, the reduced cost is close to the lower limit $2/L$. The computational efforts discussed here mainly refer to the computation cost of operating the convolutional kernels on features in the original dense block and our LDB, excluding the calculation cost of concatenation in dense block and the calculation cost of the feature addition after conversion to sum in LDB. The main reasons are twofold. On one hand, the calculation cost of the applied sum and concatenation operations on extracted features is far less than that of the convolution kernels, which is not in one order of magnitude. On the other hand, the calculation cost of the concatenation operator is not easy to measure, so we mainly focus on computing the cost of convolution kernels.

## 4. Compressed DenseNet for the Lightweight Character Recognition

We present the lightweight character recognition network based on the designed LDB. We design the architecture of compressed DenseNet with an up-sampling block, termed CDenseNet-U. CDenseNet-U mainly has two parts, namely, convolutional layer and transcription layer. Notice that the traditional CNNs usually extend the depth of the network by stacking more convolutional and down-sampling layers. But the size of stacked features cannot be reduced forever, and the down-sampling layer may also cause some useful feature information loss. To consummate these deficiencies, we use deconvolution to define an up-sampling block based on the original CNNs, similarly as [52], which can help to extend the depth and restore lost feature information to a certain extent. To improve the model efficiency of our CDenseNet-U, we use the depth-wise separable convolution to replace the original convolution operation. Moreover, we design two different convolution operation group for better compression. One is an operation group that includes BN, ReLU and depth separable convolution. The other one is by adding a convolution with kernel size of 1*1, and a scale factor $t$ for reducing the channels of input, based on the former convolution group. Note that the convolution units with 1*1 convolution are deployed in the first layer of LDB, transition layers and Conv2, since the input features of these layers usually have large number of channels. Besides, we use the convolution operation with stride 2 to replace the pooling as a down-sampling strategy to prevent the feature information being lost and make the parameters learnable at the same time in the transition layer. The structure of our CDenseNet-U framework has two parts, i.e., LDB based encoding module and up-sampling block. Figure 4 shows the convolutional part of CDenseNet-U.

**LDB encoding module.** This feature encoding module includes a convolution layer, three LDBs and two transition layers. The first convolution layer, i.e., Conv1, is mainly applied to extract shallow features and also plays a role of down-sampling when encountering with large-size features. It is worth noting that this module mainly operates the convolution and down-sampling on the input images and outputs dense encoding features.

**Up-sampling block.** This up-sampling block optimizes the dense features further from the LDB encoding module and outputs the up-sampling features as the input of the transcription layer. Up-sampling block is the core of our proposed CDenseNet-U, which contains the operations of up-sampling, two lightweight dense blocks, a transition layer and a convolution layer. The most important part of this block is up-sampling, and a good up-sampling way can effectively restore the lost feature information, which is widely-used in the image restoration. In addition, the up-sampling block is helpful to extend the depth of network for delivering deeper features. On the basis of the original image pixels, we use the deconvolution [49] to construct our up-sampling block. Note that for the study on CNNs, the deconvolution usually refers to the inverse process of the convolution operation. Similar to the convolution, deconvolution also involves the multiplication and the addition operation. In addition, the deconvolution can be used to up-sample the CNN features to the resolution of original images.

**Transcription layer.** This transcription layer is mainly used to transform the prediction of each frame into the final label sequence, which includes the operations of soft-max and CTC. The soft-max function is employed to output the predictions of the learned features from convolution parts, and CTC can transform the predictions into the final label sequence. In our CDenseNet-U framework, CTC needs to input data of each column of a picture containing text as a sequence and outputs the corresponding characters.

## 5. Experimental Results and Analysis

In this section, we evaluate CDenseNet-U on two character recognition tasks: (1) recognizing the texts in images [38-40]; (2) recognizing the handwritten character images. For the first task, we compare the results of our CDenseNet-U with those of several related deep models, where the CPUs and GPUs of all the evaluated methods in experiments are Xeon E3 1230 and 1080 Ti respectively and the applied convolution architectures are based on the framework of Caffe [31]. A large-scale synthesis Chinese String dataset [30] is used for the evaluations. For the second task, we compare the recognition results with some popular methods on MNIST [15] and HASY [56] for handwritten character recognition. It is noteworthy to point out that CTC is required in the first task of character recognition, but is not needed in the second task of digits recognition. Because the second task only needs to perform the single character recognition, we use the convolution parts of CDenseNet-U to extract deep features and then use the soft-max function [3] as the classifier to predict the labels of samples.

### 5.1. Handwritten Character Recognition

#### 5.1.1. Experiments on MNIST

We first evaluate each deep framework for recognizing the handwritten digits based on images using the popular MNIST database [15]. Note that MNIST is a widely-used handwritten digit dataset, where the goal is to classify the images with 28×28 pixel as one of the 10 digital class. The MNIST handwritten digit dataset contains 60,000 training samples and 10,000 testing samples. MNIST can evaluate the feature learning ability of a learning algorithm. In Fig.5, we show some image examples of the MNIST dataset.

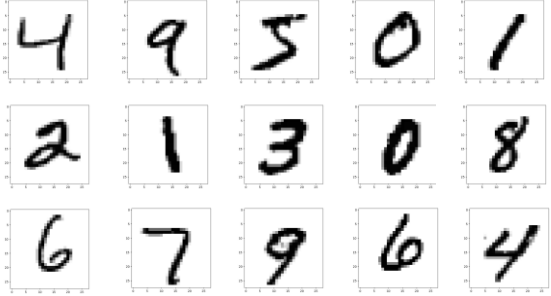

Fig. 5. Illustration of some image examples in MNIST.

| Evaluated Frameworks | Accuracy (%) |
|---|---|
| Deep L2-SVM | 99.13% |
| Maxout Network | 99.06% |
| BinaryConnect | 98.71% |
| PCANet-1 | 99.38% |
| gcForest | 99.26% |
| Simple CNN with BaikalCMA loss | 99.47% |
| **CDenseNet-U (ours)** | **99.64**% |

Table 1. Comparison of handwritten digit recognition results on the MNIST database.

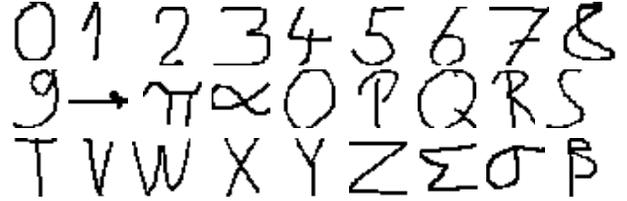

Fig. 6. Illustration of some image examples in HASY.

| Evaluated Frameworks | Accuracy (%) |
|---|---|
| Random Forest | 62.4% |
| MLP | 62.2% |
| LDA | 46.8% |
| CNN-3 | 78.4% |
| CNN-4 | 80.5% |
| CNN-4A | 81.0% |
| CNN-3+ displacement features | 78.8% |
| CNN-4+ displacement features | 81.4% |
| CNN-4A+ displacement features | 82.3% |
| **CDenseNet-U (ours)** | **84.8**% |

Table 2. Comparison of handwritten symbol recognition results on the HASY database.

**Implementation details.** For MNIST, the batch size is set to 128 and the epoch size is 200 in this experiment. The initial learning rate is set to 0.001, which will be adjusted to 0.0001 at interval between 50 and 100, and to 0.00001 after 100 epochs. Note that we add a fully-connected layer after the last convolutional layer in our CDenseNet-U so that the output features can be transformed into the required form of softmax. To prevent the overfitting, we add four dropout layers after Conv1, LDB3, Conv2 and the fully-connected layer, and set the value to 0.5 in CDenseNet-U. In this study, the performance of CDenseNet-U are compared with those of six popular deep models, including Deep L2-SVM [41], Max-out Network [42], BinaryConnect [43], PCANet-1 [44], gcForest [45] and Simple CNN with BaikalCMA loss.

**Recognition results.** The handwritten digit recognition results in terms of accuracy on MNIST are described in Table 1. We see that our frameworks achieve the enhanced results, and the accuracies reach 99.64%, which implies that the proposed LDBs have strong representation ability.

### 5.1.2. Experiments on HASY

We then evaluate each deep model for recognizing the handwritings of HASY. HASY is a public dataset of single symbols but more challenging, because the number of classes in HASY is more than that of MNIST and there are many similar classes in HASY, i.e., it has 168,233 instances with 369 classes. Some examples are shown in Fig.7.

**Implementation details.** For the task of handwritten classification using 10-fold cross-validation, the samples of HASY are divided into training set and test set with the proportion of 9:1, and 10 datasets that are divided into different images in the same proportion are provided. In the experiment of each dataset, the batch size is set to 64, and the epoch size is set to 20. The initial learning rate is set to 0.001 and will be adjusted to 0.0001 between 10 and 20. A fully-connected layer is added after Conv2 in CDenseNet-U. To prevent over-fitting, a dropout layer [44] is added after the fully connected layer, and the parameter value is set to 0.5. We average the convergence results on each dataset to obtain the final classification accuracy.

**Recognition results.** We compare our model with five popular deep models, i.e., random forest [53], multi-layer perceptron (MLP) [54], linear discriminant analysis (LDA) [55], CNN-3/4/4A [56] and (CNN-3/4/4A) + displacement features [57]. The random forest uses a large number of decision trees in the integrated classifier; LDA is a linear classifier; MLP is a fully- connected forward neural network; CNN-3/4/4A is a multi-layer CNN model; (CNN-3/4/4A)+ displacement features is a multi-layer CNN model by using displacement features. It can be seen from the classification results in Table 2 that our CDenseNet-U outperforms other competitors by delivering better performance.

### 5.2. Character Recognition in Images

In this section, we evaluate each deep framework model for recognizing the texts in images by using the synthetic Chinese string dataset. This string dataset is generated from Chinese corpus, including news and classical Chinese, by changing fonts, sizes, gray levels, blurring, perspective and stretching, which is made by following the procedures in

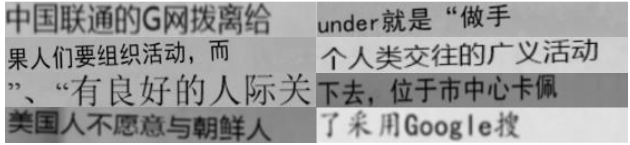

Figure 7. Some image examples in the Chinese string dataset.

| Evaluated Frameworks | Accuracy (%) |
|---|---|
| Inception-bn-res-blstm | 92.00% |
| ResNet-res-blstm | 91.00% |
| DenseNet-res-blstm | 96.50% |
| DenseNet-no-blstm | 97.00% |
| DenseNet-sum-blstm-full-res-blstm | 98.05% |
| DenseNet-no-blstm-vertical-feature | 98.16% |
| DenseNet | --- |
| DenseNet-UB (Bilinear) | 98.89% |
| DenseNet-UB (Deconvolution) | 99.27% |
| **CDenseNet-U (ours)** | **99.45**% |

Table 3. Comparison of text recognition results of evaluated deep models on the synthetic Chinese string database.

[30]. The dictionary has about 5990 characters, including Chinese, punctuation, English and numbers. Each sample is fixed to 10 characters, and those characters are randomly intercepted from the corpus. The resolution of text pictures is unified into 280×32. A total of about 3 million 600 thousand images are generated, which are divided into a training set and a test set according to 9:1. Fig.7 illustrates some image examples of the Chinese string dataset.

**Implementation details.** We use the stochastic gradient descent (SGD) [28] for training and take Tensorflow [32] and Keras as our experiment architectures. The training of the deep network is implemented on TITAN Xp. The batch size of CDenseNet-U is set to 32. The epoch size is 10. The initial learning rate is set to 0.001, which will be adjusted at each epoch with algorithm of 0.005*0.4**epoch, where "**" denotes the power calculation. The weight decay is set to 0.0001. In our CDenseNet-U, we add a dropout layer [27] after the Conv2 and set the dropout rate to 0.2 to prevent overfitting. We use the value of test loss as a metric, and the training stops when the loss values do not descend. The weights are kept when the training of each epoch finishes.

**Character recognition results.** Table 3 describes the recognition result of each model, where "Accuracy" refers to the correct proportion of the whole string and statistics on test set. For the compared models and our model, the recognition results are based on the frameworks of CRNN/DenseNet+CTC. Specifically, the frameworks with suffix "res-blstm" denotes the models with blstm in the form of residuals, the frameworks with suffix "no-blstm" means that there is no LSTM layer. "DenseNet-sum-blstm-full-res-blstm" has two changes over "Densenet-res-blstm" framework: (1) the approach of combining the two lstms into blstm changes from concat to sum; (2) both layers of blstm are connected by the residual way. "DenseNet-no-blstm-vertical-feature" removes the pooling operations [26] of 1x4 to "Densenet-no-blstm" relatively. "DenseNet-UB" denotes the original DenseNet with an un-sampling block [33], where the bilinear interpolation and Deconvolution are used to construct the un-sampling block. "DenseNet" represents the original framework of the "DenseNet-UB" without up-sampling block, which is under-fitting, so we cannot give the result of this method. We see that our CDenseNet-U can achieve higher accuracies up to 99.45%, compared with other related models. The experimental results once again demonstrate that our proposed compressed dense neural network with lightweight dense blocks can obtain better results greatly reduced computation complexity. The used dataset and results of most compared methods are publicly available at https://github.com/senlinuc/caffe_ocr.

| Evaluated Frameworks | Weight size (MB) |
|---|---|
| DenseNet-UB | 90.99 |
| CDenseNet-U (channel=64) | 78.48 |
| CDenseNet-U ($t$=1/2) | 78.84 |
| CDenseNet-U ($t$=1/4) | 77.95 |
| **CDenseNet-U ($t$=1/8)** | **77.50** |

Table 4. Comparison of produced weight sizes of DenseNet-U and our CDenseNet-U with different compression factors on MNIST.

| Evaluated Frameworks | Weight size (MB) |
|---|---|
| DenseNet-UB | 45.70 |
| CDenseNet-U (channel=64) | 33.26 |
| CDenseNet-U ($t$=1/2) | 33.55 |
| CDenseNet-U ($t$=1/4) | 32.66 |
| **CDenseNet-U ($t$=1/8)** | **32.21** |

Table 5. Comparison of produced weight sizes of DenseNet-U and our CDenseNet-U with different compression factors on HASY.

## 5.3. Comparison of Weight Compression

Since the lightweight design is the main focus of this paper, we would like to present some experiments to evaluate the weight compression capability of our CDenseNet-U based on different compression scale factors $t$. In this simulation, MNIST and HASY are evaluated. Tables 4 and 5 show the comparison of the weight sizes of DenseNet-UB and our CDenseNet-U on MNIST and HASY, respectively. Since CDenseNet-U is defined based on DenseNet-UB with some compression strategies, and both of their frameworks are very similar except the used dense blocks and two different convolution units designed in this work. Thus, comparing the weight sizes of DenseNet-UB [33] and CDenseNet-U are fair and reasonable. In Tables 4 and 5, CDenseNet-U (channel=64) means that the compression channel of 1*1 convolution is 64 and $t$ is the compression factor. We can see that the weight sizes of CDenseNet-U under different compression factors are smaller than that of DenseNet-UB, and the weight size becomes smaller with the decrease of $t$. The results clearly verified the effectiveness of our weight compression strategies for reducing the weight size. Note that the results of our CDenseNet-U in Table 2 and 3 are got under the case that channel=64. That is, our CDenseNet-U can also obtain promising character recognition results in addition to compressing the weight size.

## 5.4. Visualization of Recognized Texts in Images

In addition to the above quantitative evaluation results, we also visualize some recognized texts in images by using our CDenseNet-U framework in Fig.8. In order to visualize the results, we utilize the CTPN to extract the key text lines from the testing images. We observe that our networks can output high-quality character recognition results. Note that for some identified sentences, there have some deviations in the text positions, which is due to the fact that CTPN has no layout analysis function making it fail to produce the accurate text alignment when detecting text lines.

# 6. Remarks on Compressed DenseNet-U

## 6.1. About the First Convolution Layer

The first convolution layer (i.e., Conv1) in our models can apply different functions according to different inputs. For input images with small size, Conv1 mainly extracts the shallow features. But when facing the large-size images, the Conv1 usually set its stride as 2 to down-sample the original inputs. In this case, Conv1 will play the same role as the transition layer, while the difference is that the kernel size of the Conv1 in the whole framework is 5*5 for bigger receptive field, not 3*3 in other convolution operations. In theory, we can obtain better results with deeper networks and input images with larger sizes that contain more feature information. But due to the limit of computing power and resources, we usually need to adopt the down-sampling strategy to features with very large size. In fact, we can also magnify the original input images by up-sampling such as transpose convolution. Under these circumstances, we can further extend the depth of our proposed network and the final performance is also expected to be improved.

## 6.2. Strategies to Extend the Depth of Network

We propose a compressed dense network CDenseNet-U. In fact, CDenseNet-U also presents two strategies to help expand the depth of network. Specifically, CDenseNet-U uses the deconvolution as the up-sampling block to enable the network to add more convolution layers. By this way, we can also expand the depth of network by stacking more up-sampling blocks or including more LDBs in the block of up-sampling. In fact, we can also integrate the two depth extending methods of our CDenseNet-U. As such, we can potentially obtain stronger feature leaning ability.

## 6.3. Discussion on Up-sampling

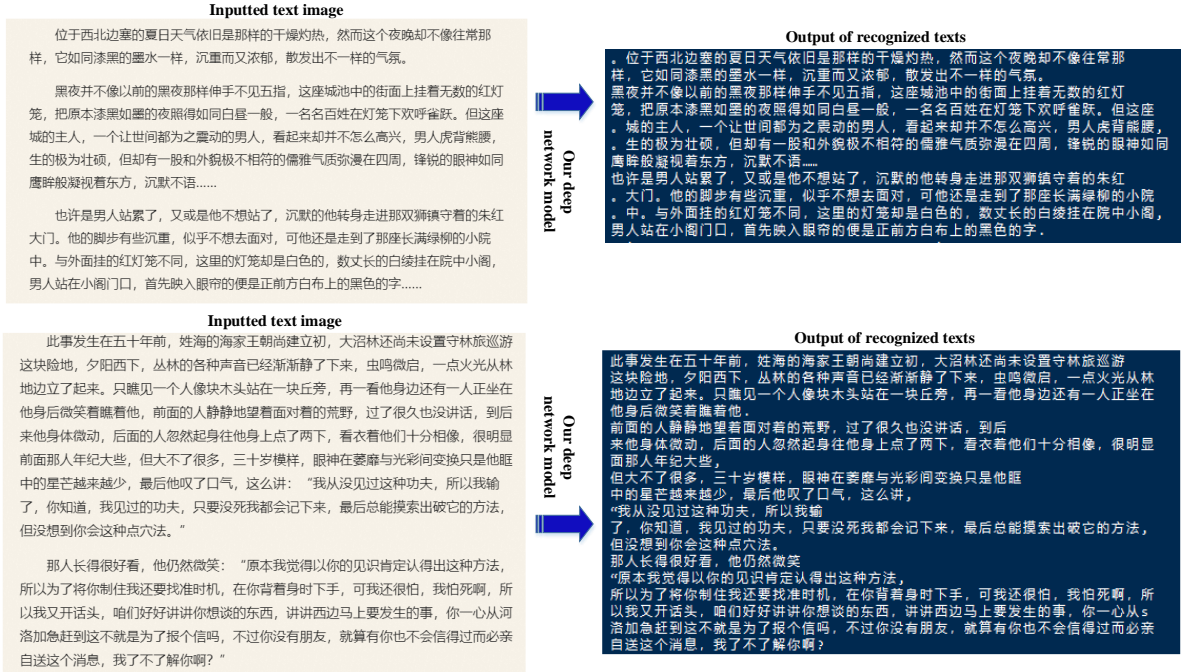

Figure 9. Illustration of some Chinese recognition results using our proposed CDenseNet-U framework.

To define the up-sampling block in our CDenseNet-U, different up-sampling methods can be applied, for instance traditional interpolation, edge-based interpolation, bilinear interpolation [47], region-based interpolation [48] and deconvolution methods [49]. But choosing an appropriate interpolation algorithm to insert new elements between the pixels is important. By comparing with the deconvolution, other interpolation methods mentioned-above are based on the fixed algorithm to sample the images or features, so the results are different due to different algorithms. But a bad up-sampling method fails to effectively recover the feature information. As such, we choose the deconvolution as our up-sampling method, because it can learn the parameters adaptively in the training process.

### 6.4. Discussion on Application Areas

By the theoretical analysis and experimental verification, we conclude that our LDB can clearly reduce the weight size and also obtain more promising performance in the character recognition tasks at the same time. In addition, LDB is applicable to any deep networks that the original dense block can be used. Due to the fact that DenseNet has a very wide range of application areas and the dense blocks have also been used by many existing deep network models, we believe that we can replace the original dense block to construct different LDB based deep network models to deal with different recognition problems, including the other computer vision tasks. Moreover, since CDenseNet-U is a full-convolutional network, we can similarly infer that our

CDenseNet-U will obtain promising results potentially in the task of object segmentation [34-35].

### 7. Conclusion and Future Work

We have proposed a new convolutional block termed lightweight dense block (LDB), and proposed a lightweight character recognition network termed CDensenet-U. The framework of CDensenet-U employs several strategies for weight compression, including proposing LDB and using the depth separable convolution and adding a convolution with the kernel size of 1*1 and a scale factor for reducing the channels of input. Specifically, LDB can clearly reduce the computing cost and weight size to $(1/L, 2/L)$, compared with the dense blocks. CDensenet-U is constructed based on the designed LDBs. To restore feature information and extend the depth of network, CDenseNet-U constructs an up-sampling block using several LDBs and deconvolution.

We examined the performance of our CDenseNet-U for both character string and handwritten character recognition. From the investigated cases, we see that CDenseNet-U can clearly reduce the weight size during the model training, and can also obtain promising character recognition results, compared with related deep models. Although promising result has been obtained by CDenseNet-U, there are some issues that are still worthy of exploring. For example, more effective ways to reduce the computing cost and weight size of the dense blocks, and retain the important feature information are still to be studied. Besides, it is also very interesting to deploy the lightweight character recognition network in mobile devices, e.g., mobile phone, in future.